\documentclass[a4paper,fleqn]{cas-dc}

\usepackage[numbers]{natbib}
\usepackage[ruled]{algorithm2e}

\def\tsc#1{\csdef{#1}{\textsc{\lowercase{#1}}\xspace}}
\tsc{WGM}
\tsc{QE}
\tsc{EP}
\tsc{PMS}
\tsc{BEC}
\tsc{DE}

\begin{document}
\let\WriteBookmarks\relax
\def\floatpagepagefraction{1}
\def\textpagefraction{.001}
\shorttitle{}
\shortauthors{Mingjie Sun et~al.}

\title [mode = title]{Adaptive ROI Generation for Video Object Segmentation Using Reinforcement Learning}                      

\author[1]{Mingjie Sun}[orcid=0000-0002-3697-7927]

\ead{mingjie.sun18@xjtlu.edu.com}

\author[1]{Jimin Xiao}[]
\cormark[1]

\ead{jimin.xiao@xjtlu.edu.com}

\author[1]{Eng Gee Lim}

\author[1]{Yanchu Xie}

\author[2]{Jiashi Feng}

\address[1]{Xi'an Jiaotong-Liverpool University, Suzhou, China}
\address[2]{National University of Singapore, Singapore}

\cortext[cor1]{Corresponding author}

\begin{abstract}
In this paper, we aim to tackle the task of semi-supervised video object segmentation across a sequence of frames where only the ground-truth segmentation of the first frame is provided. The challenges lie in how to online update the segmentation model initialized from the first frame adaptively and accurately, even in presence of multiple confusing instances or large object motion. The existing approaches rely on selecting the region of interest for model update, which however, is rough and inflexible, leading to performance degradation. To overcome this limitation, we propose a novel approach which utilizes reinforcement learning to select optimal adaptation areas for each frame, based on the historical segmentation information. The RL model learns to take optimal actions to adjust the region of interest inferred from the previous frame for online model updating. To speed up the model adaption, we further design a novel multi-branch tree based exploration method to fast select the best state action pairs. Our experiments show that our work improves the state-of-the-art of the mean region similarity on DAVIS 2016 dataset to 87.1\%.
\end{abstract}


\maketitle

\section{Introduction}

\begin{figure}
	\centering
	\includegraphics[width=0.95 \linewidth]{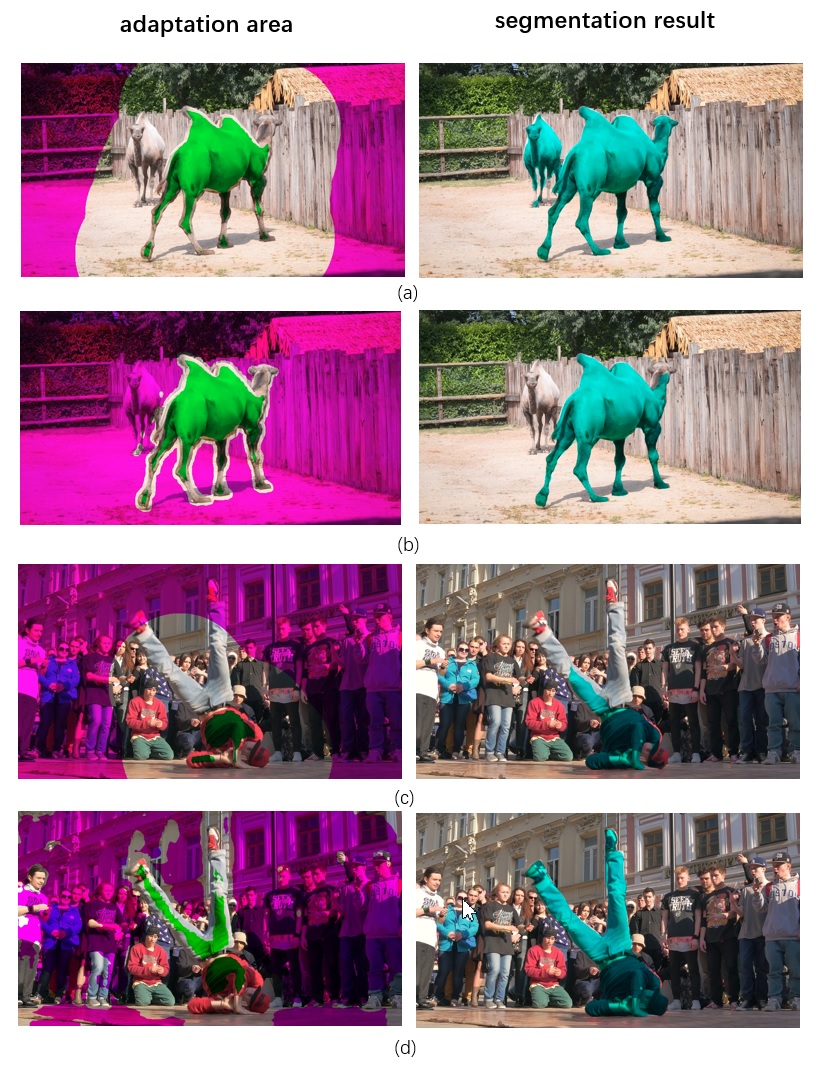}\\
	\caption{Different adaptation areas lead to different frame segmentation results. The segmentation models are the same before online adaptation. In the left column, the area in green is the adaptation area for foreground and the area in purple is the adaptation area for background.}
	\label{intuitive}
\end{figure}

Video object segmentation (VOS)~\cite{wang2017Hierarchically,li2002joint,luiten2018premvos,maninis2017video,bao2018cnn,cheng2018fast} is a fundamental problem in the computer vision field with many applications including video editing ~\cite{asai2004digital,hua2003ave}, video surveillance~\cite{brutzer2011evaluation,wang2013intelligent}, and scene understanding~\cite{cordts2016cityscapes,li2009towards}. The objective of single target video object segmentation is to label each pixel as foreground or background in a given frame of a video sequence. Labeling these pixels, however, is difficult due to background clutter, illumination change, motion blur, deformation, and so on.

There are many kinds of segmentation methods including semi-supervised video object segmentation~\cite{perazzi2017learning,chen2018blazingly,xiao2018monet}, unsupervised video object segmentation~\cite{koh2017primary,tokmakov2017learning,jain2017fusionseg,cheng2017segflow,li2018instance}, interactive video segmentation~\cite{chen2018blazingly,wang2005interactive,price2009livecut,wang2014touchcut,jain2016click,shankar2015video,pont2015semi} and so on. 
For semi-supervised segmentation, the ground-truth annotation for each pixel of the first frame is provided. The segmentation model identifies pixels of foreground and background for the following frames. 

Most recent approaches addressing semi-supervised vid-\\eo object segmentation are based on convolutional neural networks. In particular, the one-shot video object segmentation (OSVOS), introduced by Cacells et al. ~\cite{caelles2017one} has achieved great success. The core idea is to fine-tune an ImageNet~\cite{deng2009imagenet} pre-trained ConvNet on the video object segmentation data-set, such as DAVIS ~\cite{perazzi2016benchmark,pont20172017}, to allow the segmentation model to find general objects in a frame at first. Then, the first frame of the inference video sequence will be used to fine-tune the segmentation model such that it can rapidly focus on the specified target object instance in the first frame. Despite the promising result obtained using OSVOS, such a method is challenged by the cases where the appearance of the target object changes dramatically in the video sequence and there may exist multiple confusing instances of similar appearance. OSVOS only learns the target object appearance from the first frame of a video sequence. It cannot adapt to the target object appearance variation when deformation occurs or the camera rotates. 

Inspired by the fact that the online adaption has achieved great progress on video object tracking at bounding box level\\~\cite{wu2019instance,liu2020l1,han2019robust,choi2017attentional,bibi2016target,danelljan2016adaptive,xie2018correlation}, some works start to deploy the online adaption to video object segmentation. For instance, online adaptation video object segmentation (OnAVOS)~\cite{voigtlaender2017online} proposes an online adaptive video object segmentation meth-\\od which enables to update the segmentation model during inference time. However, model drift may interrupt the model online adaptation and lead to performance drop.

We observe that, for online VOS model update, how to identify to the region of interest (ROI) for model adaptation is critical. In particular, as can be observed in Figure \ref{intuitive}, if we select the adaptation area in a rough way, the segmentation model after adaptation performs very poor when there are multiple similar objects in the frame, especially, the distraction object is very close to the target. As such selecting the optimal adaptation area in a more sophisticated and flexible way is very significant for video object segmentation. 

To tackle this problem, we formalize VOS as a conditional decision-making process where two reinforcement lea-\\rning (RL) agents are employed to infer and adjust the ROI for adaption for both foreground and background in a flexible way. At each frame, the RL agent outputs actions to tune the ROI. Provided with such regions, the VOS model can be updated to be more specific and discriminating for the instance in the current frame. 

To select the optimal adaptation area for video object segmentation, a set of features of different adaptation areas of the current frames will be fed into the RL model. Then, the RL model will select the best action, and to choose the most suitable adaptation area for the current frame.  As a result, the segmentation model can obtain an accurate segmentation result. Though RL method is promising for region identification, it is notoriously slow to optimize the agent. In this work, to speed up the RL agent training, we propose a multi-branch tree based policy search method where possible action state pairs are organized in a tree structure. 

To sum up, this paper has three main contributions:
\begin{itemize}
	\item
	We observe the importance of identifying the ROI for VOS model adaptation and make the attempt to mine optimal adaptation areas for online adaptation in video object segmentation. Specifically, we deploy the ``actor-critic'' RL framework to train the agent for generating the adaptation areas. 
	
	\item
	Both VOS with online adaptation and the RL model training are computational demanding processes. To speed up the RL training, we further design a novel
	multi-branch tree based exploration method to fast select
	the best state action pairs.
	
	\item
	The proposed approach has been validated on DAVIS 2016, SegTrack V2 and Youtube-Object dataset. New state-of-the-art result of mean region similarity is obtained for the DAVIS 2016 dataset, which is 87.1\%.

\end{itemize}

\section{Related Work}
\label{sec2}
\subsection{Video Object Segmentation}

Recently, with the popularity of deep neural network, various deep learning based video object segmentation models have been proposed. These existing approaches can be classified into three different categories, including unsupervised methods, weakly supervised methods and semi-super-\\vised methods. Unsupervised methods and weakly supervised methods are more difficult than semi-supervised methods because no pixel-level annotations are available. Utilizing motion information and detecting the primary object in a frame are two common ways to address
this problem. In~\cite{tokmakov2017learning}, Pavel et al. combine the object appearance information and the motion information together successfully and achieve a good performance. In~\cite{song2018pyramid}, Song et al. propose to use concatenated pyramid dilated convolution features and dramatically improve the final accuracy. 

Semi-supervised video object segmentation task, where the pixel-level annotation of the first video frame is available, is also an extensively studied task. The most common approach for this task is to pre-train a general segmentation network. Then, the network is fine-tuned using the first frame annotation to enable the network focusing on the particular object in the frame~\cite{caelles2017one}. In order to adapt to object appearance variation, a novel approach to update the segmentation network during test time is proposed in~\cite{voigtlaender2017online}. In ~\cite{luiten2018premvos}, another method is proposed where proposals will be generated first, and then they will be merged into accurate and temporally consistent pixel-wise object tracks. In~\cite{bao2018cnn}, this task is viewed as a spatio-temporal Markov random filed problem, and ConvNet is utilized to encode the dependencies among pixels. To overcome the shortage of training data, it is proposed to use static images to generate additional training samples in~\cite{wug2018fast}. In~\cite{park2018meta}, Everingham et al. attempt to utilize the part-based tracking method to generate bounding box and segmentation for each part. In~\cite{xiao2019online}, Xiao et al. apply meta leaning to video segmentation and dramatically speeds up the segmentation process.   

Different from the existing works, we formulate the selection of ROI for online segmentation adaptation as a Markov decision process and utilize the RL to address this problem. 

\subsection{Deep Reinforcement Learning}
RL algorithm learns to achieve a complex objective from past experience. ``actor-critic''~\cite{konda2000actor} is a popular RL framework that inherits several previous RL frameworks including deep Q-learning~\cite{mnih2013playing} and policy gradient~\cite{sutton2000policy} which are valued-based and policy-based strategies, respectively. 

RL has been applied to many areas of computer vision, in particular for visual object tracking at bounding box level. In~\cite{yun2017action}, Yun et al. use RL to choose sequential actions to move the bounding box step by step from the original object location in the previous frame to correct location. In~\cite{huang2017learning}, Huang et al. propose a novel RL based model which can choose the most suitable number of deep convolutional layers according to the current frame complexity, which dramatically reduces the running time. In~\cite{dong2018hyperparameter}, Dong et al. propose to deploy the RL to choose the optimal hyper-parameter for correlation filter. 

In pixel-level video object segmentation, to the best of our knowledge, however, there is only one attempt to apply RL to this task. Han et al. establish a novel RL framework which can choose the optimal object box and the context box~\cite{han2018reinforcement}. This work is motivated by the observation that, for an identical segmentation model, different object boxes and context boxes generate different segmentation masks. Thus, it is natural to utilize RL to choose an optimal object-context box pair to achieve the best segmentation result.

Different from the existing work~\cite{han2018reinforcement}, our work aims to utilize the RL to choose the optimal ROI to update the segmentation network during test time.

\begin{figure*}
	\centering
	\includegraphics[width=0.95 \linewidth]{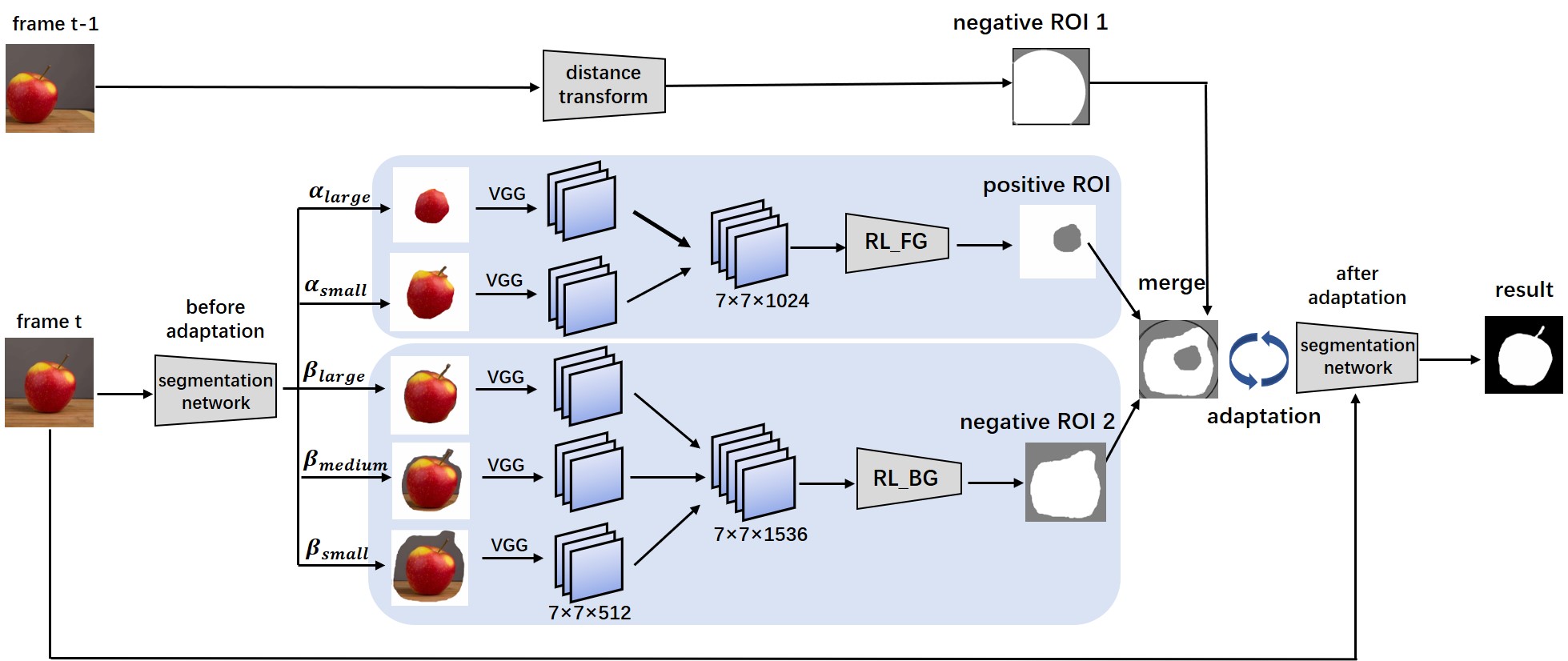}
	\caption{The network architecture of our work consists of two RL models. One is to choose the adaptation area for foreground, and another is to choose the adaptation area for background. We use pre-trained VGG19 model to extract the feature of each ROI area of the current frame, and then feed the combined feature into the RL model as the state. Finally, the RL model will choose the best adaptation area and update the segmentation model using the chosen adaptation area.}
	
	\label{framework}
\end{figure*}

\section{Our Approach}
\label{sec3}

\subsection{Overview}

In general, the main objective of our work is to utilize RL to improve the performance of video object segmentation with online adaptation. Different from the existing  approaches to select the adaptation ROI in a rough way, our work utilizes RL to choose the optimal adaptation ROI for each individual frame to avoid model drift. In other words, different frames will own its particular standard to choose adaptation ROI. The overview diagram of our proposed met-\\hod is described in Figure \ref{framework}. 

\subsubsection{VOS with Online Adaptation}

Our work is built on the top of VOS model with online adaptation. Before finally segmenting the current frame $F_{t}$  of a video sequence, a part of pixels in $ F_{t} $ will be used to update the current segmentation model to adapt to the change of the foreground and background. These pixels consist of positive ROI regarded as foreground, denoted as $S_p$, and the negative ROI regarded as background, denoted as $S_n$. In OnAVOS~\cite{voigtlaender2017online}, $ S_{p} $ and $ S_{n} $ are chosen according to two fixed thresholds including $ T_{p} $ and $ T_{n} $. In order to determine $ S_{p} $, $ F_{t} $ will be fed into the segmentation model and a temporary (before online adaptation) probability map $ M_{f} $, whose size is the same as the $ F_{t} $, will be generated. In $ M_{f} $, for a pixel $ i $, $ M_{f}(i) $ refers to the probability that pixel $i$ belongs to the foreground of $ F_{t} $. Then, $ S_{p} $ will be chosen according to $ T_{p} $, as follows:
\begin{equation} 
S_{p} =
\{i|i\in F_{t}, M_{f}(i)>T_{p}  \}.
\label{fg_threshold}
\end{equation}
Negative ROI $S_{n}$ is decided by the distance to the positive ROI $S_p$. $ S_{n} $ are the pixels far away from $S_p$, as follows:
\begin{equation}
S_n =
\{i|i\in F_{t}, distance(i)>T_{n}  \},
\label{distance}
\end{equation}
where $distance(i)$ refers to the distance of pixel $i$ to $S_{p}$.

\subsubsection{RL Based Online Adaptation}

Existing VOS methods with online adaptation, i.e., OnAVOS~\cite{voigtlaender2017online}, adopt a fixed standard to select the adaptation area, including $ T_{p} $ and $ T_{n} $, where different characteristics of each frame are not considered. To address this problem, we use flexible thresholds, $t_p$ and $t_n$, to choose the adaptation ROI. We build two RL models to choose the most suitable $ t_{p} $ and $ t_{n} $ for each frame. 

Our RL framework includes state $ s\in S $, threshold selection action $ a^{p}\in A^{p} $ to determine the value of $ t_{p} $ and threshold selection action $ a^{n}\in A^{n} $ to determine the value of $ t_{n} $, state transition function $ {s}'=T (s,a^{p},a^{n}) $ and the reward function $ g(s,a^p,a^n) $. Given a frame $ F_{t} $ of one sequence, first of all, $ F_{t} $ will be fed into the segmentation network to obtain a temporary probability map $ M_{f} $. A set of ROIs can be obtained according to the candidate thresholds. In our work, we need 5 ROIs where the probability value is greater than $\alpha_{large}$, $\alpha_{small}$, $\beta_{large}$, $\beta_{medium}$ and $\beta_{small}$, respectively. Note that $\beta_{micro}$ is ignored to diminish the complexity of the state. The first two ROIs are used for the RL model to choose $ t_{p} $ while the last three ROIs are used for RL model to choose $ t_{n} $. In other words, the possible values of $ t_{p} $ can be $\alpha_{large}$ or $\alpha_{small}$, and the possible values of $ t_{n} $ can be $\beta_{large}$, $\beta_{medium}$, $\beta_{small}$ or $\beta_{micro}$. After $ t_{p} $ and $ t_{n} $ are determined, pixels with probability values greater than $ t_{p} $ will be viewed as the adaptation ROI for foreground, as follows:
\begin{equation} 
S_{p} =
\{i|i\in F_{t}, M_{f}(i)>t_{p}  \},
\label{fg_threshold}
\end{equation}
while these pixels with probability value less than $ t_{n} $, combined with the negative pixels chosen by distance using (\ref{distance}), will be regarded as the adaptation ROI for background, as follows:
\begin{equation} 
\begin{split}
S_{n} = \{i|i\in F_{t}, distance(i)>T_{n}  \}  \\ \cup \{i|i\in F_{t}, M_{f}(i)<t_{n} \}.
\label{ourSn}
\end{split}
\end{equation}
$S_p$ and $S_n$ are used to update the current segmentation model while other pixels are ignored. After the segmentation model has been updated, $ F_{t} $ will be fed into the new segmentation network and the final segmentation result is obtained. 

The pseudo-code of our algorithm is described in Algorithm~\ref{algorithm1}.

\begin{algorithm}
	\caption{RL Based Video Object Segmentation}
	\KwIn{\\Ground-truth of the first frame $gt(1)$\\
		Sequence length $L$\\
		Distance threshold $T_n$\\
		Segmentation network $Seg\_Net$\\
		Pretrained VGG netwrok $VGG$\\
		RL model to choose positive threshold $RL_{p}$\\
		RL model to choose negative threshold $RL_{n}$\\
	}
	\KwOut{Segmentaion result of Frame $t$ $O_t$}
	
	\vspace{2ex}
	
	Fine-tune $Seg\_Net$ on $F_1$\\
	$last_{-}mask \leftarrow gt(1)$\\
	\For{$t=2$ to $L$}
	{
		Obtain 2 RL states using (\ref{state1}) and (\ref{state2}), respectively.\\
		Feed the states into the RL models and achieve the optimal threshold $t_p$ and $t_n$.\\
		Obtain positive ROI and negative ROI using (\ref{fg_threshold})(\ref{ourSn}), respectively. 
		
		Update $Seg\_Net$ on $F_t$ using $S_p$ and $S_n$.\\
		$O_{t} \leftarrow forward(Seg\_Net, F_t) > 0.5$\\
		$last_{-}mask \leftarrow O_t$
		
	}
	\label{algorithm1}
\end{algorithm}


\subsection{Agent Action}

\begin{figure}
	\begin{center}
		\includegraphics[width=0.94\linewidth]{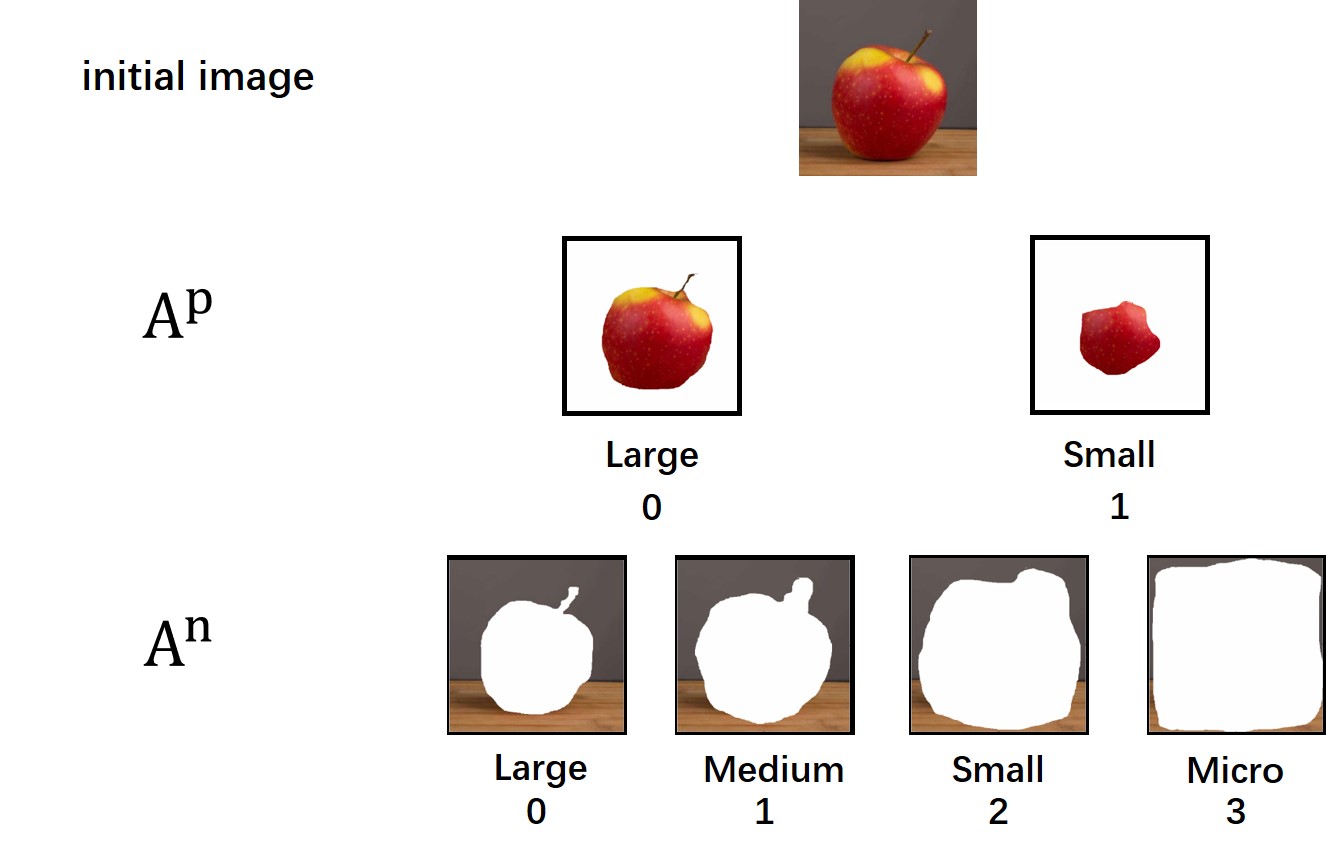}
	\end{center}
	\caption{Foreground adaptation ROI selecting action set $A^p$ and background adaptation ROI selecting action set $A^n$.}
	\label{actions}
\end{figure}

The framework of our work consists of two RL models, including one to choose $ t_{p} $ and another to choose $ t_{n} $, as shown in Figure \ref{actions}. The action set $ A^{p} $ used for the first RL model to choose $ S_{p} $ contains 2 candidate thresholds: $\alpha_{large}$ and $\alpha_{small}$. As the difference of the ROI areas with different
$t_p$ is not very big, we only set two candidate thresholds
for the action set $A^p$. The action set $ A^{n} $ used for the second RL model to choose $ S_{n} $ contains 4 candidate thresholds: $\beta_{large}$, $\beta_{medium}$, $\beta_{small}$ and $\beta_{micro}$. As the size of $S_n$ is much larger than that of $S_p$, we set more candidate thresholds for $t_n$.

In terms of the design of the candidate values of $ t_{p} $, in OnAVOS~\cite{voigtlaender2017online}, $ T_{p} $ is 0.97 and it is fixed for all frames. This threshold is very safe and conservative because it should work in any situation, especially for some frames with very bad segmentation result. In fact, for some frames with good segmentation result, the value of $ t_{p} $ ought to be much lower, so more correct pixels of the object can be used to update the model and improve the final segmentation result. As the difference of the size of the adaptation ROIs with different $ t_{p} $ is not very huge, we only set two candidate thresholds for the action set $ A^{p} $, which can lower the difficulty of training the RL model. 

In terms of the design of the candidate values of $ t_{n} $, in OnAVOS~\cite{voigtlaender2017online}, $ S_{n} $ are chosen according to the distance to the target object rather than the value of the $ M_{f} $. If a pixel is far away from the target, it will be viewed as $ S_{n} $ . Similarly, the selection of $ S_{n} $ in OnAVOS~\cite{voigtlaender2017online} is very conservative as it works for almost all frames. In our work, we still keep the adaptation ROI which are chosen by the distance, and then try to add more areas to include more correct adaptation ROI for background. The additional area is chosen by the value of $ M_{f} $. Finally, we combine these two parts of areas together as the final area of $ S_{n} $ using (\ref{ourSn}).
As the size of the area for $ S_{n} $ is much larger than the area of $ S_{p} $, we set more candidate thresholds for $ t_{n} $. 

In this way, after training, these two RL models can cho-\\ose the best action and achieve the optimal thresholds $ t_{p} $ and $ t_{n} $.

\subsection{State and Reward}

\noindent

The state $ s $ is the input of the RL model. As we have two RL models, we need two sets of states for two models. In general, the state $ s $ is a feature map combined by the feature maps of different candidate adaptation ROIs where the probability value of all pixels in a certain adaptation ROI is less than a certain threshold.

First, given a certain frame $ F_{t} $, we feed $ F_{t} $ into the segmentation model before test time adaptation and generate the temporary probability map $ M_{f} $. Then, we generate 5 ROIs with 5 different candidate thresholds according to the value of $M_f$. The ROIs where the probability values of all pixels in this area are greater than $\alpha_{large}$ and $\alpha_{small}$, receptively, are combined as the state of the RL model to choose $ t_{p} $, as follows,
\begin{equation} 
\begin{split}
state_{p} = feature(\{i|i\in F_{t}, M_{f}(i)>\alpha_{large}  \}) \\+ feature(\{i|i\in F_{t}, M_{f}(i)>\alpha_{small}  \}).
\label{state1}
\end{split}
\end{equation}
The ROIs where the probability values of all pixels in this area are greater than $\beta_{large}$,  $\beta_{medium}$ and  $\beta_{small}$, receptively, are combined as the state of the RL model to choose $ t_{n} $, as follows, 
\begin{equation} 
\begin{split}
state_{n} = feature(\{i|i\in F_{t}, M_{f}(i)>\beta_{large}  \}) \\+ feature(\{i|i\in F_{t}, M_{f}(i)>\beta_{medium}  \}) \\+ feature(\{i|i\in F_{t}, M_{f}(i)>\beta_{small}  \}).
\end{split}
\label{state2}
\end{equation}

Note that $\beta_{micro}$ is ignored in (\ref{state2}) to diminish the complexity of $state_n$, as $feature(\{i|i\in F_{t}, M_{f}(i)>\beta_{small}\})$ already provides sufficient information to help the RL model make the decision. We use the VGG model~\cite{simonyan2014very}, pre-trained on the ImageNet classification dataset~\cite{deng2009imagenet}, to extract the features of these ROIs first. Then, we concatenate these features together as the state of the RL model. We use the first 5 convolutional blocks of the VGG19 model which results in a feature size of  $\mathbb{R}^{7\times7\times512} $ for one ROI. For the RL model to choose $ t_{p} $, the features of two ROIs will be concatenated to generate the final state $ s_{t_{p}} \in \mathbb{R}^{7\times7\times1024} $. For the RL model to choose $ t_{n} $, the features of three ROIs will be concatenated to generate the final state $ s_{t_{n}} \in \mathbb{R}^{7\times7\times1536} $.
Finally, states $ s_{t_{p}} $ and $ s_{t_{n}} $ will be fed into the corresponding RL model and result in the actions to choose the optimal thresholds $ t_{p} $ and $ t_{n} $.

The reward function is defined as $ r_{t}=g(s_{t},a_{p},a_n) $ which reflects the performance of the final segmentation result of each frame in the video sequence:

\begin{equation} g(s_t,a_t,a_n)=\left\{\begin{matrix}
IOU+1 & IOU>0.1 \\ 
-1 & IOU<=0.1 
\end{matrix}\right.
\label{reward},
\end{equation}
where $IOU$ indicates the intersection-over-union (IOU) between the prediction and the ground-truth, which reflect the quality of the predicted segmentation. 

\subsection{Training in Actor-Critic Framework}

In our work, we adopt the ``actor-critic'' framework~\cite{konda2000actor} for RL training. In general, one ``actor-critic'' framework consists of two roles including an ``actor'' role to generate an action and a ``critic'' role to measure how good this action is. In this work, we need to select the optimal adaptation ROIs for both foreground and background separately. Therefore, we need two ``actor-critic'' model pairs, including  one ``actor-critic'' pair for foreground, and another pair for background. Four individual RL models are deployed in total. 

In our ``actor-critic'' framework, given a current frame $F_t$, the first step is to feed the state into the ``actor'' network and generate an action $a$, which is to choose the optimal adaptation ROIs. The corresponding reward $r_t$ will also be obtained after conducting this action. $r_t$ is decided by the IOU of the segmentation result according to (\ref{reward}). 

In the training process, after the forward process, the ``critic'' network will be updated first in the valued-based way, as follows:
\begin{equation}
w = w{}' + \alpha*\delta_t\nabla_{w{}'} V_{w{}'}(s_t), 
\label{criticupdate}
\end{equation}
where
\begin{equation}
\delta_t =  r_t+ \gamma*V_{w{}'}(s_{t+1})-V_{w{}'}(s_t).
\label{tderror}
\end{equation}
In (\ref{criticupdate}) and (\ref{tderror}), $w$ and $w{}'$ indicate the weight of the ``critic'' model after and before update.  $\alpha$ is the learning rate of the ``critic'' model. $\delta_t$ is the TD error which indicates the difference of the actual score and the predicted score. $V_{w{}'}(s_t)$ refers to the accumulated reward of state $s_t$ which is predicted by the ``critic'' model before update. $\gamma$ refers to the discount factor.

After the ``critic'' model has been updated, the ``actor'' model will be updated in a policy-based way, as follows:
\begin{equation} 
\theta  = \theta{}' + \beta * \nabla (log\pi_{\theta{}'} (s_{t},a_{t})) * A(s_{t},a_{t}), 
\label{actor}
\end{equation}
where $A(s,a)$ refers to the advantage function, and $A(s_{t},a_{t})=\delta_t$ according to (\ref{tderror}), $\theta$ and $\theta{}'$ indicate the weight of the ``actor'' model after and before update. $\beta$ is the learning rate of the ``actor'' model. Policy function $\pi(s,a)$ is a network whose input is the state $s$ and a certain action $a$, and output is the probability of selection action $a$ in state $s$. 

In this way, when training the RL models, our ``actor-critic'' framework can avoid the shortage of value-based and policy-based methods. Instead of waiting until the end of the episode, our RL models can be updated at each step, which dramatically reduces the training time but maintains the RL training stability. 

\section{Implementation Details}

\subsection{Train the Segmentation Network}

The proposed method trains the segmentation network follows the strategy of~\cite{caelles2017one} and~\cite{voigtlaender2017online}. The first step is to train a ImageNet pre-trained network on a pixel-level annotated dataset such as PASCAL VOC~\cite{everingham2010pascal}. In the second step, we use the DAVIS video dataset~\cite{perazzi2016benchmark} to train the network so that the network is able to adapt to this dataset. We also fine-tune the network on the first frame of DAVIS test videos whose ground-truth annotation is provided. In this way, the segmentation network is well trained and can achieve a good result before test time adaptation.

\subsection{Train the RL Model}
Before training the RL model, the related data need to be stored in advance to accelerate the training process, which will be described in section 4.3. In terms of the training of the RL models, specifically, we divide all video sequences in DAVIS 2016 training set into video clips with the fixed number of frames. A video clip includes 10/5 consecutive frames is used as a sample for the RL model to select foreground/background ROI. Using stored clips for training dramatically reduces the training time of the RL model.

We randomly select 20 clips as a batch for training the RL models. At the beginning of the training, the learning rate $\alpha$ for ``actor'' model is 1e-5, and the leaning rate $\beta$ for ``critic'' model is 5e-5. The learning rate decreases gradually during the training, and it decreases by 1\% for each 200 iterations. The discount rate $\gamma$ for the reward is 0.9. In terms of values for candidate threshold $t_p$ and $t_n$, we found $\alpha_{large}$=0.97, $\alpha_{small}$=0.7, $\beta_{large}$=0.4, $\beta_{medium}$=0.2, $\beta_{small}$=0.1 and $\beta_{micro}$=\\0.01 works well through cross validation. The training of our RL models takes about 3 days on a NVIDIA GTX 1080 Ti GPU and a 12 Core Intel i7-8700K CPU@3.7GHz. 

\subsection{Accelerating RL Training}
Segmentation with online adaptation is slow because the segmentation model should be updated for each frame. Meanwhile, RL training itself is also slow. Training the RL model heavily relies on a large number of attempts for different actions. Normally, to generate a well trained RL model, the model should be trained with more than one million iterations. If the running time for each training process is too long, the total time is unbearable. Thus, it is impossible to train the RL model in a regular way.

\begin{figure}
	\begin{center}
		\includegraphics[width=0.94\linewidth]{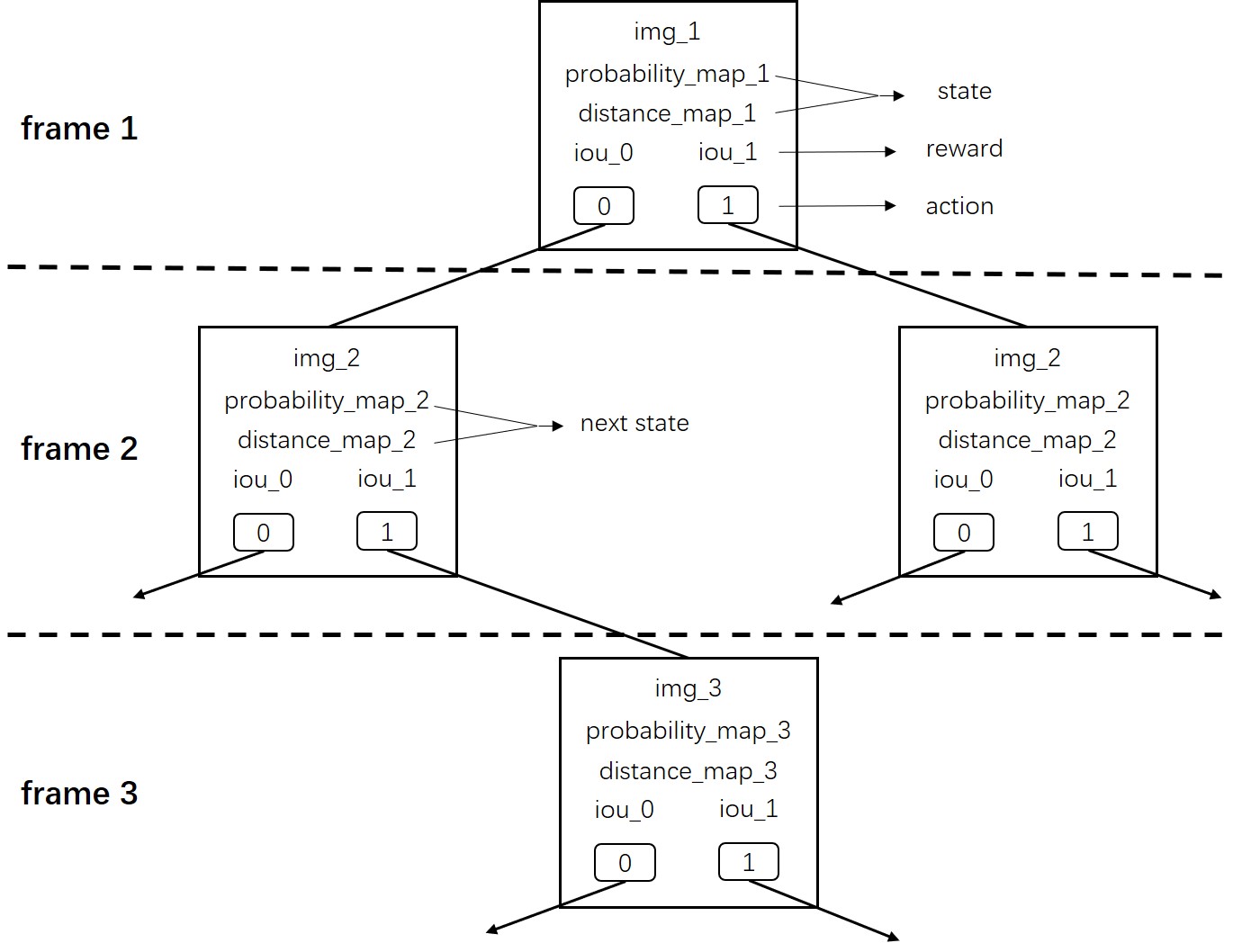}
	\end{center}
	\caption{The data structure used to restore the related data to accelerate the RL model training. For RL model to select the $t_n$, distance map is required for the training, while the training for the RL mode to selected $t_p$ does not need it.}
	\label{accelerate}
\end{figure}

To address this issue, inspired by the idea of sacrificing space to improve efficiency, we propose a novel multi-branch tree structure, as shown in Figure \ref{accelerate}, to store all possible segmentation results using different adaptation ROIs into a repository in advance. Training an ``actor-critic'' framework needs 4 types of information including the action, the reward, the states before and after the action. For each node in the multi-branch tree, a corresponding directory is generated. The image of the frame, the temporary probability map and the IOU of the segmentation result after executing a certain action are stored as files. All possible actions are stored as a links to next layer of nodes. In this way, the image and the probability map are used to generate the state. The IOU value is used to generate the reward. Finally, this repository will be organized in a multi-branch tree structure, whose stored data will be used to update the RL model during the process of training using the method described in section 3.4.

\begin{table*}[]
	\begin{center}
		\begin{tabular}{|p{70pt}|p{60pt}<{\centering}|p{60pt}<{\centering}|p{70pt}<{\centering}|p{70pt}<{\centering}|p{40pt}<{\centering}|}
			\hline
			Method&  DAVIS-16 $J_m$ &  DAVIS-16 $F_m$& SegTrack  V2 $J_m$  & Youtube Objs $J_m$  &t(s) \\
			\hline\hline
			PReMVOS~\cite{luiten2018premvos}         &84.9          &\textbf{88.6}            &-                &-  &30    \\
			OnAVOS~\cite{voigtlaender2017online}     &\underline{85.7}          &84.8            &66.7             &77.4   &13  \\
			CINM~\cite{bao2018cnn}                   &83.4          &85.0            &77.1             &\underline{78.4}   &120   \\
			Lucid~\cite{khoreva2017lucid}            &83.7          &-               &76.8             &76.2   & 40    \\
			MSK~\cite{perazzi2017learning}           &79.7          &75.4            &72.1             &75.6  &12    \\
			OSVOS~\cite{caelles2017one}              &79.8          &80.6            &65.4             &78.3  &10 \\
			STV~\cite{wang2017super}                 &73.6          &-               &\textbf{78.1}             &-   &-   \\
			ObjFlow~\cite{tsai2016video}             &68.0          &-               &74.1             &77.6    &- \\
			OURS                                     &\textbf{87.1} &\underline{86.1}            &\underline{77.5}             &\textbf{79.5}  & 14  \\
			\hline
		\end{tabular}
	\end{center}
	\caption{Quantitative comparison with other methods on the DAVIS 2016, SegTrack V2 and Youtube-Object dataset. For $J_m$ and $F_m$, the method wiht the best performance is bold, and the method with the second best performance is marked with underline.}
	\label{compare}
	
\end{table*}

\section{Experiments}

\subsection{Experiment Setup}

Our method is evaluated on three widely-used datasets including the DAVIS 2016 dataset~\cite{perazzi2016benchmark}, Youtube-Object data-set~\cite{prest2012learning} and Segtrack V2 dataset~\cite{li2013video}. DAVIS 2016 dataset consists of 50 high quality video sequences and 3,455 frames, spanning multiple occurrences of common video object segmentation challenges such as occlusions, motion-blur and appearance changes. 30 video sequences of DAVIS 2016 are used for training, and 20 video sequences are used for testing. In DAVIS 2016, in each video sequence, only a single object instance is annotated. DAVIS 2017 dataset~\cite{pont20172017} extends the DAVIS 2016 dataset where multiple objects, rather than only one object, are annotated in each frame. As our method targets for single instance segmentation, we only do the experiment on DAVIS 2016. In Youtube-Object, there are 155 video sequences and a total of 570,000 frames. These video sequences are divided into 10 classes. Training set and testing set are not separated in Youtube-Object dataset so it is only used for testing. In SegTrack V2 dataset, there are 14 video sequences with more occlusion than appearance changes compared with Youtube-Object dataset. 

We evaluate our method following the approach proposed in~\cite{perazzi2016benchmark}. The adopted evaluation metrics include region similarity $J$ and contour accuracy $F$. The region similarity is calculated as $J=\left | \frac{m\cap gt}{m\cup gt} \right |$ by the intersection-over-union between the predicted segmentation $m$ and the ground-truth $gt$. The contour accuracy is defined as $F=\frac{2P_cR_c}{P_c+R_c}$, which indicates the trade-off between counter-based precision $P_c$ and recall $R_c$.

\begin{table}[]
	\begin{center}
		\begin{tabular}{|p{120pt}<{\centering} |p{80pt}<{\centering}|}
			\hline
			Method &  DAVIS 2016 \\
			\hline\hline
			WO adaptation &80.3 $\pm$ 0.4     \\
			foreground adaptation &  82.1 $\pm$ 0.5    \\
			background adaptation & 85.3 $\pm$ 0.5     \\
			full adaptation &  86.5 $\pm$ 0.4    \\
			full adaptation + CRF & \textbf{87.1} $\pm$ 0.4     \\
			\hline\hline
			OnAVOS~\cite{voigtlaender2017online} WO CRF &  84.6 $\pm$ 0.6  \\
			OnAVOS~\cite{voigtlaender2017online}&  85.7 $\pm$ 0.6 \\
			CINM~\cite{bao2018cnn} & 84.2 \\  
			\hline
		\end{tabular}
	\end{center}
	\caption{Ablation study on the contribution of individual RL model for the DAVIS 2016 dataset, measured by the mean region similarity $J_m$. WO indicates without.}
	\label{ablationDAVIS}
\end{table}

\subsection{Comparison with State-of-the-arts}
In this section, we will compare our proposed work with other state-of-the-art semi-supervised video object segmentation methods, including  PReMVOS~\cite{luiten2018premvos}, OnAVOS~\cite{voigtlaender2017online}, CINM~\cite{bao2018cnn}, LucidTracker~\cite{khoreva2017lucid}, MSK~\cite{perazzi2017learning}, OSVOS~\cite{caelles2017one}, STV~\cite{wang2017super}, ObjectFlow~\cite{tsai2016video}. Note that OSVOS-S~\cite{maninis2017video} is not included in the comparison list as it utilizes additional dataset for training. 

Table \ref{compare} summarizes the quantitative results of recent methods on the DAVIS 2016 validation set consisting of 20 videos. The top 3 performing methods have been highlighted with different colors. It can be observed that our work has achieved outstanding result under both mean region similarity $J_m$ and the mean contour accuracy $F_m$. Especially on mean region similarity $J_m$, our method achieves the best result which outperforms any existing state-of-the-art methods. Compared with the most competitive and related method OnAVOS~\cite{voigtlaender2017online}, our method improves the mean region similarity $J_m$ to 87.1\%. It should be noted that the gain over ~\cite{voigtlaender2017online} is solely due to the fact that better ROIs are obtained for online adaptation using RL. Note that, same to~\cite{marki2016bilateral}, according to the randomness of the segmentation network, the final accuracy may fluctuate around 0.4\%. The proposed mean region similarity $J_m$ is the average value from experiments of 10 times. 

\begin{figure*}
	\begin{center}
		\includegraphics[width=1\linewidth]{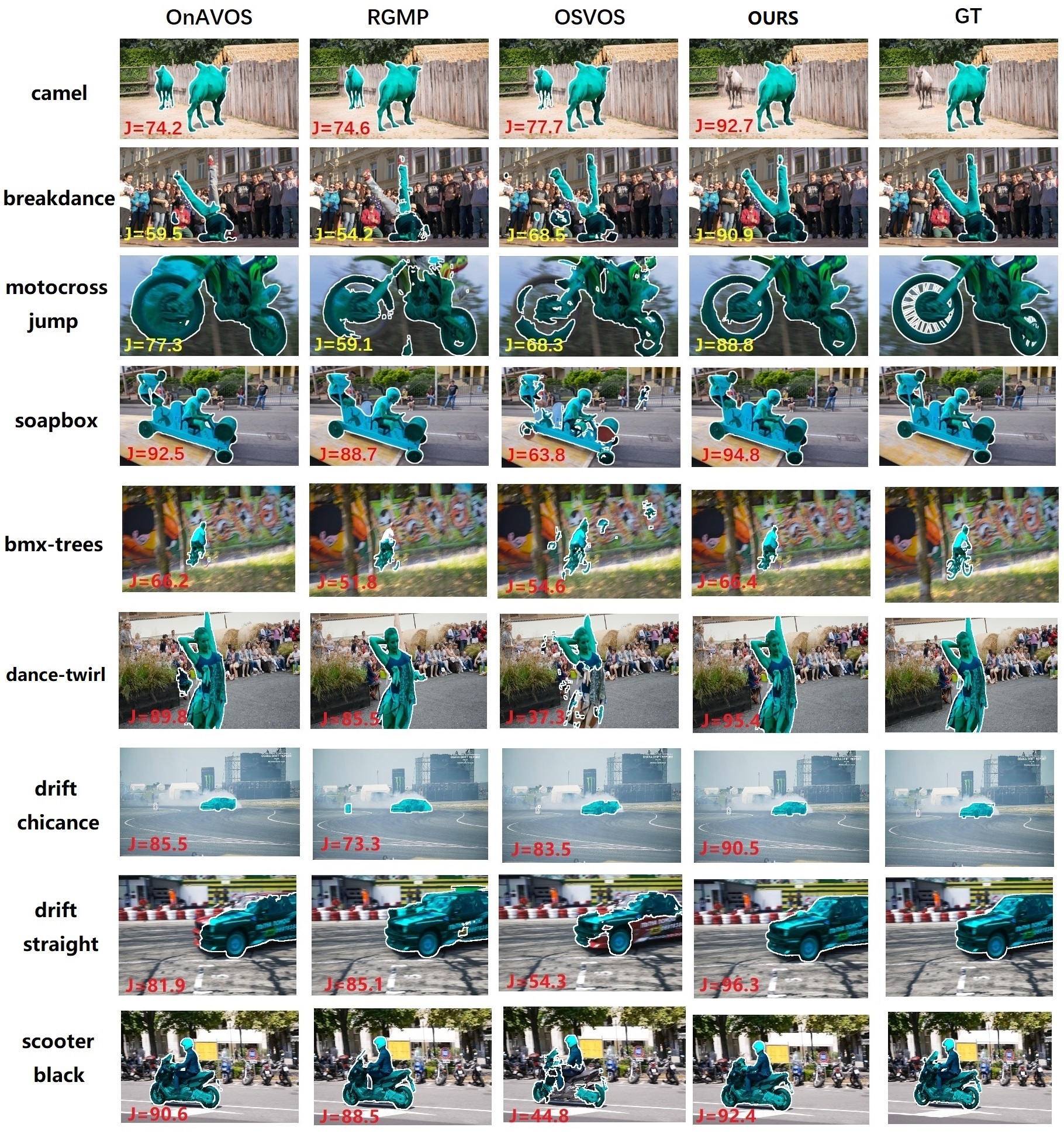}
	\end{center}
	\caption{Visualization of segmentation masks and the mean region similarity $J_m$ for different methods on the DAVIS 2016 dataset.}
	\label{visualization}
\end{figure*}

Figure \ref{visualization} shows the qualitative segmentation masks for different methods. As can be observed, our method performs better on videos with significant appearance change for the target object, for instance, the camel and breakdance video sequences. Especially when multiple similar objects are close to each other, e.g., the camel sequence, our method has the ability to distinguish the target object from other similar objects successfully. 

On SegTrack V2 and Youtube-Object datasets, as training set and evaluation set are not split, all video sequences are used for evaluation. From Table \ref{compare}, we can observe that our method also performs well on both datasets. Compared with OnAVOS~\cite{voigtlaender2017online}, our method improve the mean region similarity $J_m$ by 10.8\% on SegTrack V2 dataset, which demonstrates the effectiveness of our RL models to choose the online adaptation ROIs. In addtion, our approach also improves the mean region similarity $J_m$ by 2.1\% on Youtube-Object dataset. This result can show the robustness of our method on different evaluation datasets.

In addition, we compare our run time with other state-of-the-art methods, and the result is also reported in Table \ref{compare}. Despite the fact that we improve the mean region similarity $J_m$ of our baseline method with online adaptation~\cite{voigtlaender2017online} by 10.8\% on SegTrack V2 dataset and 1.4\% on Davis 2016 dataset, the run time of our method is only about 1 second longer than~\cite{voigtlaender2017online}, which demonstrates the efficiency of our method. 

\begin{table}[]
	\begin{center}
		\begin{tabular}{|p{120pt}<{\centering}|p{80pt}<{\centering}|}
			\hline
			Method & SegTrackV2\\
			\hline\hline
			WO adaptation  &61.4 $\pm$ 0.6    \\
			foreground adaptation  &66.2 $\pm$ 0.5   \\
			background adaptation   &73.2 $\pm$ 0.5   \\
			full adaptation  &76.6 $\pm$ 0.5   \\
			full adaptation + CRF  & \textbf{77.5} $\pm$ 0.5    \\
			\hline\hline
			OnAVOS~\cite{voigtlaender2017online} WO CRF  & 64.9 $\pm$ 0.6 \\
			OnAVOS~\cite{voigtlaender2017online} &66.7 $\pm$ 0.6\\
			CINM~\cite{bao2018cnn}  &77.1\\  
			\hline
		\end{tabular}
	\end{center}
	\caption{Ablation study on the contribution of individual RL model for the SegTrack V2 dataset, measured by the mean region similarity $J_m$. WO indicates without.}
	\label{ablationSeg}
\end{table}

\subsection{Ablation studies}

In this section, we conduct four ablation studies on our method using the testing video sequences of DAVIS 2016 dataset and SegTrack V2 dataset. 

We conduct the first ablation studies on the DAVIS 2016 and SegTrack V2 datasets, where parts of our method are disabled to investigate the impact of each component. In this study, we also explore the contribution of each individual RL model in our method, one of these two RL models is disabled during this ablation study respectively, and the results will be compared with the result generated by the method with full RL models. Table \ref{ablationDAVIS} shows the result of this ablation study on the DAVIS 2016 dataset. On the DAVIS 2016 dataset, when using both foreground and background adaptation, we obtain the mean region similarity $J_m$ of 86.5\% without CRF \cite{kr2011efficient}. $J_m$ of our method with full adaptation is greater than the method without any adaptation by 6.2\%, which demonstrates the effectiveness of our online adaptation approach. In addition, compared with the rough approach to choose the adaptation ROI adopted by OnAVOS~\cite{voigtlaender2017online}, before CRF, our method is also 1.9\% greater than it. After executing CRF, the gain degrades a little to 1.4\%, which demonstrates that our flexible way to choose the optimal adaptation ROI for each frame is significant for the final segmentation result. To study the individual influence of each RL model, we remove the RL model to choose the optimal adaptation ROI for background, obtaining the method \textbf{foreground adaptation}, and remove the model to choose the adaptation ROI for foreground, obtaining the method \textbf{background adaptation}. As can be observed from the Table \ref{ablationDAVIS}, using \textbf{foreground adaptation} method and \textbf{background adaptation} method obtains $J_m$ of 82.1\% and 85.3\%  on DAVIS 2016 dataset, respectively, which indicates that both RL models improve the segmentation result while the RL to choose the optimal adaptation ROI for background makes larger contribution to the final segmentation result. This observation also explains the reason why we have more threshold candidates for background than foreground. We also conduct the same ablation studies on the Segtrack V2 dataset and obtain the similar results as can be observed from the Table \ref{ablationSeg}. When using both foreground and background adaptation, we obtain $J_m$ of 61.4\% without CRF. $J_m$ of our method with full adaptation is greater than the method without any adaptation by 15.2\%. The comparison between the rough approach to choose the adaptation ROI adopted by OnAVOS~\cite{voigtlaender2017online} and the proposed method is conducted on Segtrack V2 dataset as well, before CRF. The result is that $J_m$ of the proposed method is 11.7\% greater than the baseline method, which is a dramatic improvement of the mean region similarity. After the process of CRF, the gain degrades a little to 10.8\%, which is still a huge improvement. To study the individual influence of each RL model on Segtrack V2 dataset, we also adopt the method \textbf{foreground adaptation} and the method \textbf{background adaptation}, as on the DAVIS 2016 dataset. As can be observed from the Table \ref{ablationSeg}, using \textbf{foreground adaptation} method and \textbf{background adaptation} method obtains 66.2\% and 73.2\% $J_m$ on Segtrack V2 dataset, respectively, which also indicates that both of the two RL models contribute to the improvement of segmentation result. This experiment also demonstrates the importance of the optimal adaptation ROI mining. 

\begin{table}[]
	\begin{center}
		\begin{tabular}{|p{60pt}<{\centering}|p{60pt}<{\centering}|p{70pt}<{\centering}|}
			\hline
			$t_n$ & $t_p$  & $J_m$\\
			\hline\hline
			0.4   &0.97   & 61.9 $\pm$ 0.6 \\
			0.4  &0.7  & 61.2 $\pm$ 0.6 \\
			0.2   &0.97  & 64.2 $\pm$ 0.6 \\
			0.2  &0.7  & 62.5 $\pm$ 0.6 \\
			0.1  & 0.97   & 73.7 $\pm$ 0.6 \\
			0.1  & 0.7   & 69.6 $\pm$ 0.6 \\
			0.01  & 0.97   & \textbf{83.6} $\pm$ 0.6 \\
			0.01  & 0.7   &  80.1 $\pm$ 0.6 \\
			\hline\hline
			\multicolumn{2}{|c|}{OURS}  & \textbf{87.1} $\pm$ 0.4  \\
			\hline
		\end{tabular}
	\end{center}
	\caption{Performance comparison between heuristic adaptation ROIs selection and RL-based adaptation ROIs selection, conducted on the DAVIS 2016 dataset, measured by the mean region similarity $J_m$. $t_n$ indicates the threshold to choose the adaptation ROI for background. $t_p$ indicates the threshold to choose the adaptation ROI for foreground.}
	\label{ablationThreshold}
\end{table}

The purpose of the second ablation experiment is to dem-\\onstrate that it is important to select different $t_n$ and $t_p$ for each specific frame, rather than adopting a particular fixed set of thresholds. In other words, no matter which set of $t_n$ and $t_p$ are chosen, as long as these thresholds are fixed for all frames, the final segmentation result will be worse than the result generated by adopting the optimal thresholds selected by the RL models for each specific frame. More specifically, the RL model to select the adaptation ROI for background has 4 candidate thresholds including $\{0.4, 0.2 ,0.1, 0.01\}$. The RL model to select the adaptation ROI for foreground has 2 candidate thresholds including $\{0.97, 0.7\}$. In this way, there are totally 8 possible combinations of $t_p$ and $t_n$, which are listed in Table \ref{ablationThreshold}. In this experiment, each combination of $t_n$ and $t_p$ is adopted as the fixed thresholds for the segmentation model adaptation and its corresponding result is evaluated and compared with the result generated by the proposed method. As can be observed from Table \ref{ablationThreshold}, among these combinations, the most ``conservative'' one, i.e. $t_p$=0.97 and $t_n$=0.01 performs best. Note that these values are different from the final result of OnAVOS\cite{voigtlaender2017online} because OnAVOS selects the background adaptation ROI according to the distance, rather than the value of the probability map. The performance decreases dramatically when the value of the fixed threshold gets close to 0.5 because the error of segmentation will propagate quickly when adopting these "greedy" thresholds for all frames. The highest obtained $J_m$ (83.6 \%) of the method adopting fixed thresholds, however, is still much lower than the result generated the proposed method (87.1 \%). According to the observation of this experiment, it is obvious that the improvement does result from the RL models that choose the optimal $t_n$ and $t_p$ for each individual frame, rather than the contribution of a certain fixed combination of $t_n$ and $t_p$.

The purpose of the third ablation experiment is to demonstrate that the proposed method is able to find a better adaptation ROI, compared with the baseline method. The metric to evaluate the selected adaptation ROI is the IOU between the adaptation ROI and the ground-truth. In this way, we firstly calculate the IOU between the adaptation ROI selected by the RL model and ground-truth. Then, we calculate the IOU between the adaptation ROI selected by the baseline method and the ground-truth. Finally, we compare these two IOUs and their contributions to the final final segmentation result. As can be observed from Table \ref{ablationIOU}, on average, the IOU of the proposed method is 40.9 \% and 19.4 \% higher than the baseline method, for background and foreground respectively. This result also coincides with the facts reported in Table \ref{ablationDAVIS} and Table \ref{ablationSeg} that background ROI adaptation using RL  brings more performance gain than foreground ROI adaptation. Finally, as more correct pixels are adopted to update the segmentation model in the proposed method, the segmentation model is able to properly adjust itself to the change of the target in the video. It is also the reason why the proposed method can achieve improvement on the final segmentation result. 

\begin{table*}[]
	\begin{center}
		\begin{tabular}{|p{75pt}<{\centering}|p{40pt}<{\centering}|p{45pt}<{\centering}|p{40pt}<{\centering}|p{45pt}<{\centering}|p{30pt}<{\centering}|p{35pt}<{\centering}|p{30pt}<{\centering}|p{30pt}<{\centering}|p{20pt}<{\centering}|}
			\hline
			Video & IOU-B-RL  & IOU-B-Base & IOU-F-RL  & IOU-F-Base & $J_m$-RL  & $J_m$-Base & $\Delta$IOU-B & $\Delta$IOU-F & $\Delta$$J_m$ \\
			\hline\hline
			blackswan       &99.4   & 44.6   &91.0   & 74.6  & 95.4  & 95.4  & 54.8  & 16.4  & 0    \\
			bmx-trees       &98.9   & 75.1   &39.9   & 15.6  & 55.5  & 52.5  & 23.8  & 24.3  & 3.0  \\
			breakdance      &97.4   & 55.9   &74.9   & 55.0  & 80.2  & 68.4  & 41.5  & 19.9  & 11.8 \\
			camel           &98.8   & 45.0   &89.9   & 70.8  & 93.8  & 84.0  & 53.8  & 19.1  & 9.8  \\
			car-roundabout  &99.4   & 41.8   &95.2   & 85.4  & 97.1  & 97.1  & 57.6  & 9.8   & 0    \\
			car-shadow      &99.7   & 60.1   &92.7   & 79.2  & 96.1  & 96.0  & 39.6  & 13.5  & 0    \\
			cows            &99.1   & 43.1   &91.6   & 76.9  & 94.6  & 94.6  & 56.0  & 14.7  & 0    \\
			dance-twirl     &98.1   & 59.3   &83.3   & 63.2  & 87.3  & 84.6  & 38.8  & 20.1  & 2.7  \\
			dog             &99.3   & 52.6   &92.6   & 78.9  & 95.1  & 95.1  & 46.7  & 13.7  & 0    \\
			drift-chicane   &99.6   & 72.1   &83.0   & 55.6  & 89.1  & 87.2  & 27.5  & 27.4  & 1.8  \\
			drift-straight  &99.1   & 64.4   &90.1   & 75.8  & 92.7  & 91.3  & 34.7  & 14.3  & 1.3  \\
			goat            &99.0   & 58.4   &88.8   & 73.7  & 91.2  & 91.1  & 40.6  & 15.1  & 0    \\
			horsejump-high  &98.9   & 61.8   &80.0   & 52.4  & 87.3  & 86.8  & 37.1  & 27.6  & 0.5  \\
			kite-surf       &98.3   & 72.4   &57.2   & 28.0  & 66.7  & 66.7  & 25.9  & 29.2  & 0    \\
			libby           &99.3   & 66.3   &74.9   & 45.1  & 86.1  & 86.1  & 33.0  & 29.8  & 0    \\
			motocross-jump  &93.9   & 41.2   &87.5   & 70.7  & 90.4  & 86.4  & 52.7  & 16.8  & 4.0  \\
			paragliding-launch   &97.1   & 69.1   &58.7   & 44.9  & 62.5  & 62.5  & 28.0  & 13.8  & 0    \\
			parkour         &99.4   & 67.3   &85.1   & 61.7  & 91.8  & 91.4  & 32.1  & 23.4  & 0.4  \\
			scooter-black   &98.6   & 58.2   &86.1   & 68.1  & 89.8  & 89.0  & 40.4  & 18.0  & 0.8  \\
			soapbox         &98.3   & 46.1   &86.0   & 65.4  & 90.1  & 86.2  & 52.2  & 20.6  & 3.8  \\

			\hline\hline
			mean & 98.6 & 57.7   & 81.4  & 62.0  & 86.5  & 84.6 & 40.9   & 19.4  & 1.9 \\
			\hline
		\end{tabular}
	\end{center}
	\caption{Quality comparison between the selected ROIs of the proposed method and the baseline method, as well as their influences on the final segmentation results, conducted on the DAVIS 2016 dataset, measured by the mean region similarity $J_m$. In this table, IOU indicates the IOU between the adaptation ROI selected and the ground-truth. $J_m$ indicates the mean region similarity. B refers the background and F refers to the foreground. RL refers to the RL model and Base refers to the baseline method. $\Delta$ refers the difference value between the RL model and the baseline method. All values in this table are calculated before CRF.}
	\label{ablationIOU}
\end{table*}

The fourth ablation experiment is to study the influence of different states for the RL model. The adopted state for the RL model also greatly affects the convergence difficulty of the training process as well as the final segmentation performance. We design three different states to study this factor. The first state is to feed the initial image (3 channels) and the temporary probability map (1 channel) into a train-from-scratch ConvNet without using pre-trained VGG model. This is because the input of the VGG model should be a 3-channel image. The second one is to feed the initial image (3 channels) concatenated with several mask channels to indicate different adaptation area, similarly, without the feature extracting of VGG model. The third one is the proposed one where we generate a set of images with different masks to indicate different adaptation areas. Feature of each image is extracted by the VGG model. Then, the combined feature will be fed into the RL model. When using the first two types of states, after full training, the RL model still cannot choose the optimal adaptation area for each frame, which demonstrates these states are not suitable for this task. We believe the main reason is that the pre-trained VGG model is able to extract a better feature which contains more information of the original image. Also, the first two ways to indicate the different adaptation areas are not sufficiently explicit and discriminative.

\section{Conclusion}

In this paper, we have proposed a novel method that can select the best adaptation ROI for each frame to take full advantage of the test time adaptation for video object segmentation. Two RL models are applied to choose the optimal adaptation area for foreground and background individually. Comprehensive experiments on three common benchmark datasets demonstrate the great performance of our method compared with other state-of-the-art methods. In future, we plan to replace the discrete-value actions with the continue-value actions to enable the RL model to find a more flexible way to choose the optimal adaptation ROI. 

\bibliographystyle{cas-model2-names}

\bibliography{cas-refs}

\end{document}